\documentclass[runningheads,a4paper]{llncs}

\usepackage{amssymb}
\usepackage{graphicx}

\usepackage{bm}
\usepackage{siunitx}
\usepackage{subcaption}
\usepackage{amsmath}
\usepackage{booktabs}

\usepackage{url}
\urldef{\mailsa}\path| {guanxj17@mails.,jfeng@, jzhou@}tsinghua.edu.cn|

\newcommand{\keywords}[1]{\par\addvspace\baselineskip
\noindent\keywordname\enspace\ignorespaces#1}

\begin{document}

\mainmatter  

\title{Pose-Specific 3D Fingerprint Unfolding}

\titlerunning{Pose-Specific 3D Fingerprint Unfolding}

%
%
\author{Xiongjun Guan%
\and Jianjiang Feng\and Jie Zhou}%

\institute{ Department of Automation, \\Beijing National Research Center for Information Science and Technology, \\Tsinghua University, Beijing 100084, China\\
\mailsa\\}

%


\maketitle

\begin{abstract}
In order to make 3D fingerprints compatible with traditional 2D flat fingerprints, a common practice is to unfold the 3D fingerprint into a 2D rolled fingerprint, which is then matched with the flat fingerprints by traditional 2D fingerprint recognition algorithms. The problem with this method is that there may be large elastic deformation between the unfolded rolled fingerprint and flat fingerprint, which affects the recognition rate. In this paper, we propose a pose-specific 3D fingerprint unfolding algorithm to unfold the 3D fingerprint using the same pose as the flat fingerprint. Our experiments show that the proposed unfolding algorithm improves the compatibility between 3D fingerprint and flat fingerprint and thus leads to higher genuine matching scores.
\end{abstract}

\section{Introduction}
Fingerprint is one of the most widely used biometric traits beacause it is very stable, easy to collect and highly distinctive. Up to now, 2D fingerprint images obtained by contact-based sensors are the dominating fingerprint image type \cite{maltoni2009handbook}. However, the quality of 2D fingerprint images is affected by factors such as skin deformation and skin humidity. Compared with 2D fingerprints, 3D fingerprints are not deformed, less affected by dry or wet fingers, and hygienic, since they are acquired in a non-contact manner \cite{kumar2018contactless}. Due to this advantage, a number of 3D fingerprint sensing technologies have been proposed \cite{chen20063d,zhao20113d,fatehpuria2006acquiring}.

Since existing fingerprint databases contain mainly 2D fingerprints and 2D fingerprint sensors are very popular, new 3D fingerprint sensors need to be compatible with 2D fingerprint sensors. Considering that 3D fingerprint sensors are expensive and large, a common application scenario is using a 3D sensor for enrollment and a 2D sensor for recognition. For compatibility, enrolled 3D fingerprints are usually unfolded into 2D rolled fingerprints \cite{chen20063d,zhao20113d,labati2012quality,wang2010fit,anitha2014performance,wang2010data,labati2011fast,dighade2012approach,fatehpuria2006acquiring,shafaei2009new}, which are then matched to query flat fingerprints by traditional 2D fingerprint recognition technology.

Generally, there are two types of 3D fingerprint unfolding methods: parametric method and non-parametric method. The parametric method uses a hypothetical geometric model to approximate the 3D shape of the finger. In these methods, 3D fingerprints are projected into a parametric model. Commonly used parametric models include cylinders \cite{chen20063d,zhao20113d,labati2012quality}, spheres \cite{wang2010fit,anitha2014performance}, and deformable cylinders \cite{wang2010data,labati2011fast}. In addition, there are methods to determine the parameters by analyzing the curvature of the fingerprint \cite{dighade2012approach}. Most of these methods use axially symmetric parametric models. However, since the finger is irregular, these models do not fully conform to the shape of the fingers. This will cause additional deformation when unfolding. The non-parametric method does not assume the shape of the finger. These methods first smooth the surface of the finger and then unfold it according to the surface curvature \cite{fatehpuria2006acquiring,shafaei2009new}. Because of special shape of finger, it cannot be spread out on a plane without tearing. 
In addition, the deformation of the fingerprint is related to its pressing pose, and the pose of the query fingerprint is not known in advance. Therefore, there is large deformation between the unfolded fingerprint and the query fingerprint when the 3D fingerprint is unfolded in the enrollment stage. This will result in false non-matches.

The problem of fingerprint distortion can also be solved in the recognition stage. In order to deal with the skin deformation problem, researchers have proposed various deformable registration algorithms and matching algorithms tolerant to skin distortion \cite{bazen2003fingerprint,ross2006image,cheng2013minutiae,si2017dense,cui20182}, which are however designed for 2D fingerprints. Since there is no explicit consideration of 3D finger shape during registration or matching, non-mated fingerprints may appear similar after unreasonable deformable registration, which may result in false matches.

In this paper, we propose a pose-specific 3D fingerprint unfolding algorithm to unfold the 3D fingerprint using the same pose as the flat fingerprint, with the aim of reducing deformation between unfolded fingerprint and flat fingerprint. We collected a database of mated 2D and 3D fingerprints and performed matching experiments on it. Experiments show that pose difference is a very important cause for fingerprint deformation and genuine matching score can be greatly improved by pose-specific unfolding.

\section{Proposed Method}\label{sec:method}

\begin{figure}[htb]
  \centering
    {\includegraphics[width=0.95\linewidth]{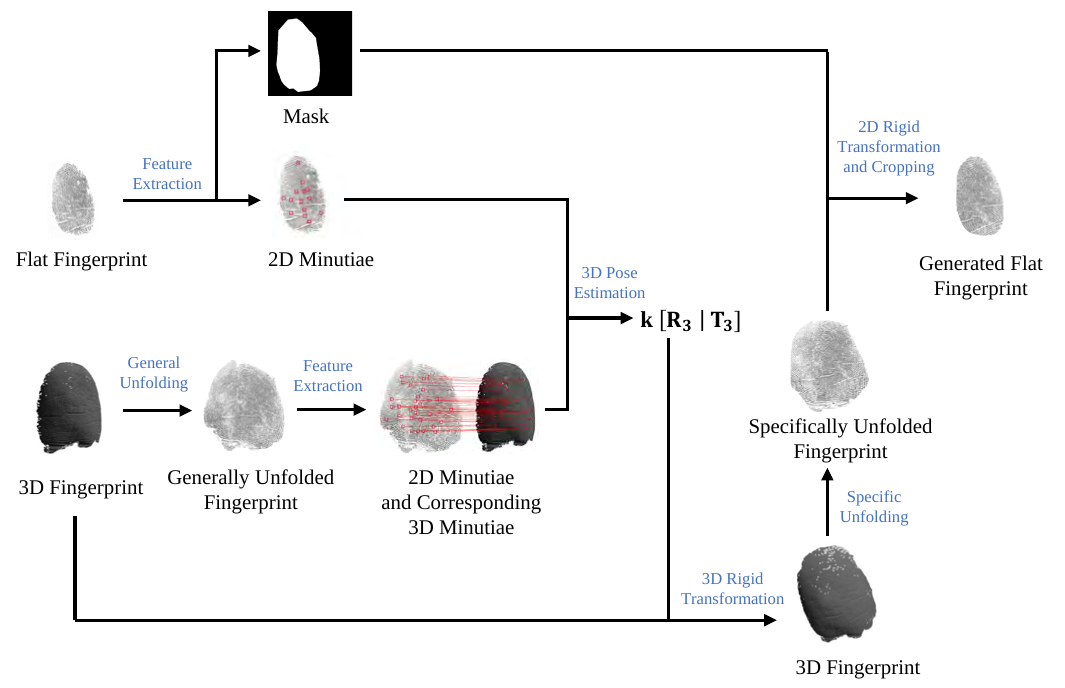}}
  \caption{Flowchart of the proposed pose-specific 3D fingerprint unfolding.}
  \label{fig:process}
\end{figure}

Fig. \ref{fig:process} shows the complete process of the proposed pose-specific 3D fingerprint unfolding algorithm. The algorithm first unfolds the 3D fingerprint using the normal pose, next estimates the specific pose of the query flat fingerprint by minutiae matching, then unfolds the 3D fingerprint using the estimated pose, and finally applies a simple 2D fingerprint registration method. The two unfolding algorithms are the same and the only difference is that the pose of the 3D fingerprint is adjusted during the second unfolding. In the following we describe the unfolding and 3D pose estimation steps, since other steps are very basic or based on existing techniques. We use ``unfolding" to represent the specific step as well as the whole algorithm. Its meaning should be clear based on the context.

\subsection{Unfolding}
General 3D model rendering techniques are not suitable for generating high quality fingerprint images from 3D point cloud for two reasons. Firstly, in 3D model rendering, the light-dark relationship is usually obtained by setting the light source, which is difficult to unify the brightness in different areas of a fingerprint image. Secondly, affected by the position of the light source, rendered bright area (fingerprint ridge) will be out of phase with real flat fingerprints. The previous method of generating fingerprints from point clouds \cite{fatehpuria2006acquiring} calculates the surface depth by fitting the smooth surface of the finger. This method may overfit and cause the ridge line to be uneven. Therefore, we propose an algorithm to get unfolded rolled fingerprints using local point sets from finger point clouds, including two steps: point cloud visualization and unfolding.

\subsubsection{Visualization}
The texture of a flat fingerprint is produced when the finger is in contact with the  sensor plane. Generally speaking, the more prominent the ridge line, the deeper the fingerprint texture. The visualization method in this work simulates this property, takes the neighborhood around each point in the point cloud as the partial surface, and calculates the depth of the point relative to the surface.

\begin{figure}[htb]
  \centering
    {\includegraphics[width=0.7\linewidth]{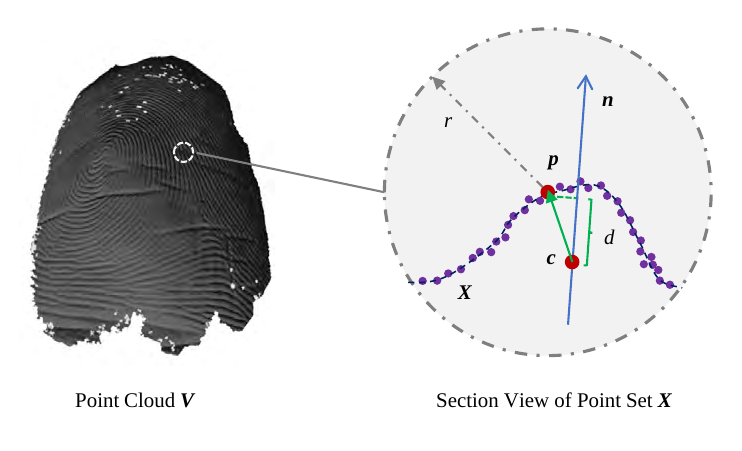}}
  \caption{Schematic diagram of the visualization method.}
  \label{fig:visual_method}
\end{figure}

The schematic diagram of point cloud visualization is shown in Fig. \ref{fig:visual_method}. 
Let $\bm{V}$ be the point set of the 3D finger. For a point $\bm{p}$ in $\bm{V}$, the neighborhood point $\bm{X}$ is defined as

\begin{equation}
  \bm{X} = \left\{\bm{x_i} \; | \; \bm{x_i}\in \bm{V} \; \text{and} \;  \left \| \bm{x_i} - \bm{p} \right\|< r \right\} \; ,
  \label{eq:def_X}
\end{equation}
where $r$ is the maximum distance from any points in $\bm{X}$ to $\bm{p}$. We set the point center as the mean of all the points in $\bm{X}$, denoted as $\bm{c}$. Let $\bm{n}$ be the normal vector of the point set $\bm{X}$ computed by principal component analysis (PCA). 
The surface depth of $\bm{p}$ relative to $\bm{X}$  is then computed as
\begin{equation}
  d = (\bm{p}-\bm{c})^{T} \bm{n} \; .
  \label{eq:def_d}
\end{equation}

\begin{figure}[htb]
  \centering
  \subcaptionbox{Thumb}
    {\includegraphics[width=0.32\linewidth]{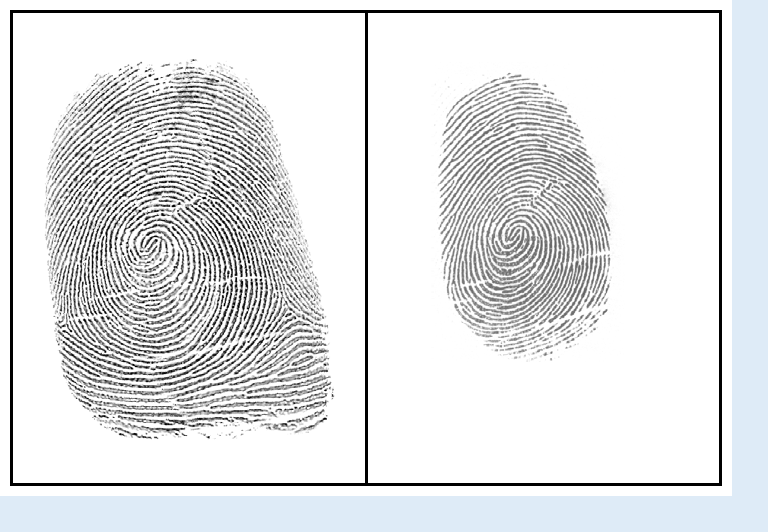}}
  \subcaptionbox{Middle Finger}
    {\includegraphics[width=0.32\linewidth]{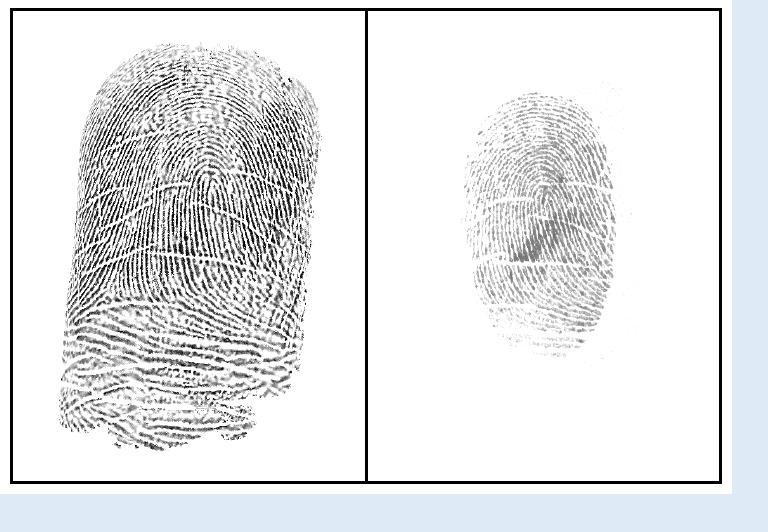}}
  \subcaptionbox{Little Finger}
    {\includegraphics[width=0.32\linewidth]{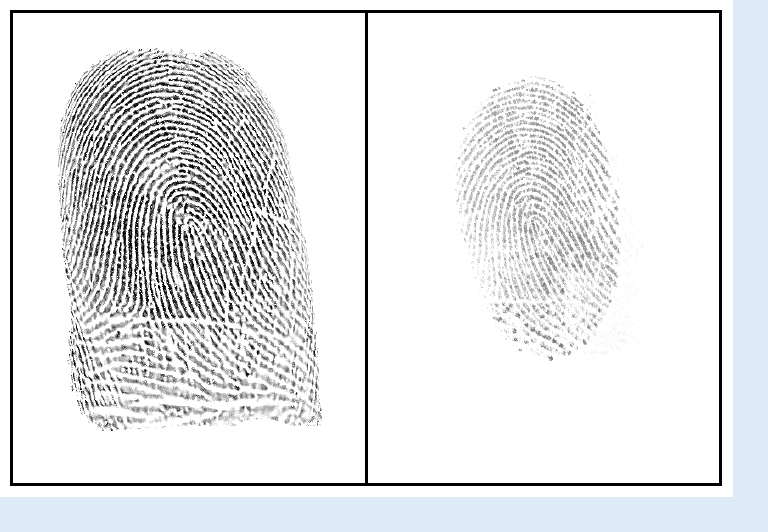}}
  \caption{Visualization results of 3D point clouds (left) and real flat fingerprints (right) in each subfigure. Image quality is similar, while the former has a larger area. There is perspective distortion in the visualization result which will be removed by unfolding step.}
  \label{fig:visual}
\end{figure}

 For each point in the 3D finger, we use the normalized surface depth $d_n$ as the pixel value which is computed as
 \begin{equation}
  d_n = \frac{d-d_{min}}{d_{max}-d_{min}} \; ,
  \label{eq:def_d_n}
\end{equation}
 and project it to the imaging plane, where $d_{max}$ and $d_{min}$ are determined from the depth $d$ of all point cloud.
 The comparison between the visualized result and the real flat fingerprint is shown in Fig. \ref{fig:visual}.

\subsubsection{Unfolding}
We use the arc length between two points in the point cloud as the coordinates of the unfolded fingerprint. Let $(x,y,z)$ be the point cloud coordinates and $(u,v,0)$ be the unfolded fingerprint coordinates. The unfolding relationship of the direction of $u$ is defined as
\begin{equation}
  \frac{\partial{u}}{\partial{x}} = \sqrt{1+(\frac{\partial{z}}{\partial{x}})^{2}}\;, \quad \frac{\partial{u}}{\partial{y}} = \sqrt{1+(\frac{\partial{z}}{\partial{y}})^{2}} \; ,
  \label{eq:uv}
\end{equation}
and the relationship of $v$ is the same.

\begin{figure}[htb]
  \centering
  \subcaptionbox{Front Pose}
    {\includegraphics[width=0.4\linewidth]{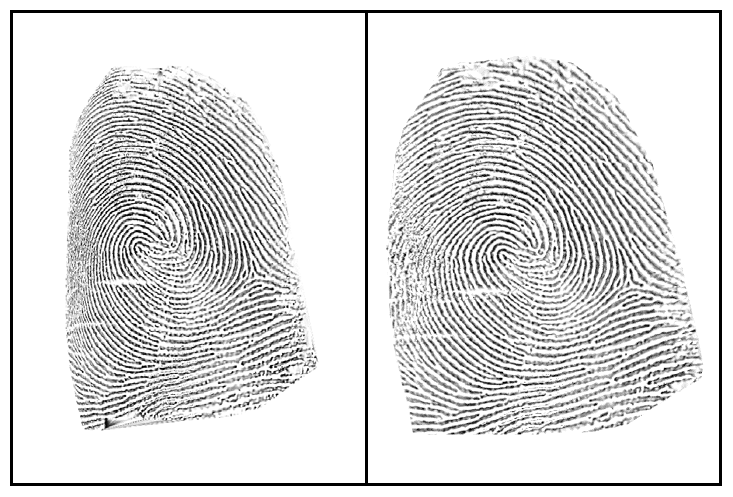}}
    \hspace{6mm}
  \subcaptionbox{Side Pose}
    {\includegraphics[width=0.4\linewidth]{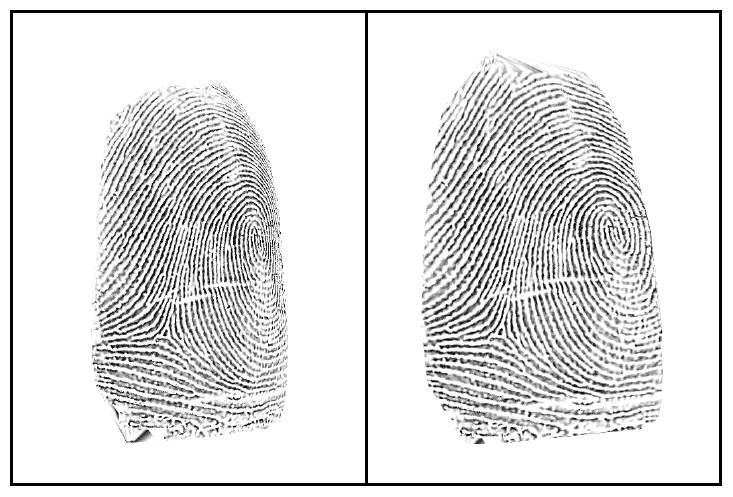}}
  \caption{Visualization results of point clouds of two fingers of different poses before (left) and after (right) unfolding.}
  \label{fig:unfold}
\end{figure}

Fig. \ref{fig:unfold} shows the comparison of the visualization results of point clouds of two fingers of different poses before and after unfolding. The area with a large tilt relative to the observation plane can be effectively flattened.

\subsection{3D Pose Estimation}
The 3D pose estimation step conducts global rigid transformation between the 3D fingerprint and the flat fingerprint. Minutiae from each unfolded rolled fingerprint and flat fingerprint are extracted using VeriFinger \cite{VeriFinger}. MCC minutia desciptor \cite{cappelli2010minutia}, which is a state-of-the-art minutiae descriptor, is used to compute similarity between all possible minutiae pairs. Then we use spectral clustering method \cite{leordeanu2005spectral} to find corresponding minutiae pairs. Since the corresponding 3D minutiae of 2D minutiae in the unfolded rolled fingerprint are known, we obtain correspondences between 3D minutiae and 2D minutiae in the flat fingerprint.

Given mated 3D/2D minutiae pairs, we estimate a 3D rigid transformation to minimize the average distance of minutia pairs under orthogonal projection. For a minutiae point $(x,y,z)$ and its projection coordinates $(u,v)$, the relationship is given as
\begin{equation}
  \begin{bmatrix} u \\ v \\ 0 \end{bmatrix} 
  = 
  \begin{bmatrix} 1 & 0 & 0\\0&1&0\\0&0&0\end{bmatrix} \; 
  (\; k \; 
  \begin{bmatrix} \bm{R_3} \; | \; \bm{t_3}  \end{bmatrix}  
  \begin{bmatrix} x \\ y \\ z \\ 1 \end{bmatrix}
  \; ) \; ,
  \label{eq:3dmatrix}
\end{equation}
where $k$ is the scaling parameter, $\bm{R_3}$ and $\bm{t_3}$ denote the rotation matrix and translate matrix of the rigid transformation in 3D. This estimation method needs at least 6 matching points.

\begin{figure}[htb]
  \centering
  \subcaptionbox{}
    {\includegraphics[width=0.3\linewidth]{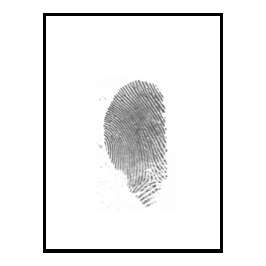}}
    \hspace{1mm}
  \subcaptionbox{}
    {\includegraphics[width=0.3\linewidth]{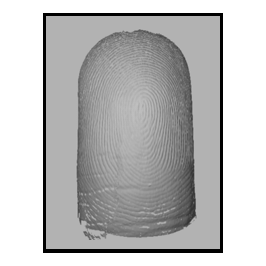}}
    \hspace{1mm}
  \subcaptionbox{}
    {\includegraphics[width=0.3\linewidth]{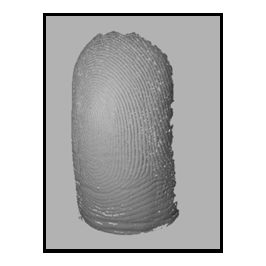}}
  \caption{Illustration of 3D pose estimation. (a) A real flat fingerprint. (b) Corresponding finger point cloud in the original pose. (c) Corresponding finger point cloud in the same pose as the flat fingerprint.}
  \label{fig:transformation}
\end{figure}

Fig. \ref{fig:transformation} shows an example of transformed 3D fingerprint using 3D pose estimation results. We run the unfolding algorithm again on the transformed 3D fingerprint to obtain the pose-specific rolled fingerprint.

\section{Experiment} \label{sec:experiment}
\subsection{Database}
Due to the lack of databases containing high quality finger point clouds and corresponding flat fingerprints, we collected a database for this study. The finger point clouds were captured by using a commercial structured light 3D scanner, while the corresponding 2D flat fingerprints at multiple poses were captured with a Frustrated Total Internal Reflection (FTIR) fingerprint sensor. 3D point clouds of 150 fingers and 1200 corresponding flat fingerprints (8 images per finger) were used in experiments. Finger data were collected from people aged 20 to 30, including 13 males and 2 females. Ten fingers were collected for each person.

\subsection{Matching Score}

\begin{figure}[htb]
  \centering
  \subcaptionbox{}
    {\includegraphics[width=0.45\linewidth]{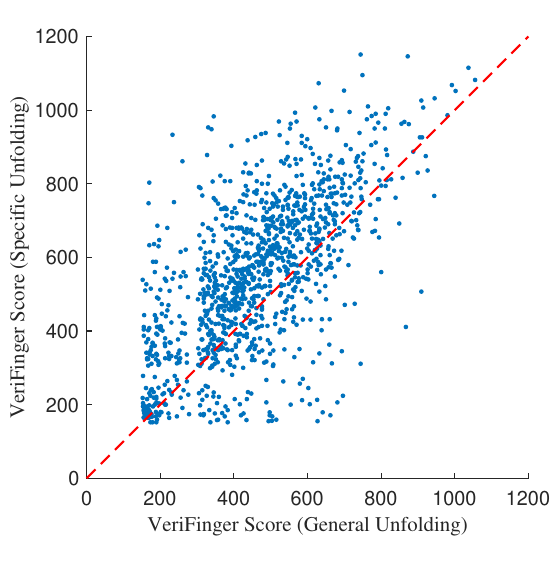}}
    \hspace{5mm}
  \subcaptionbox{}
    {\includegraphics[width=0.45\linewidth]{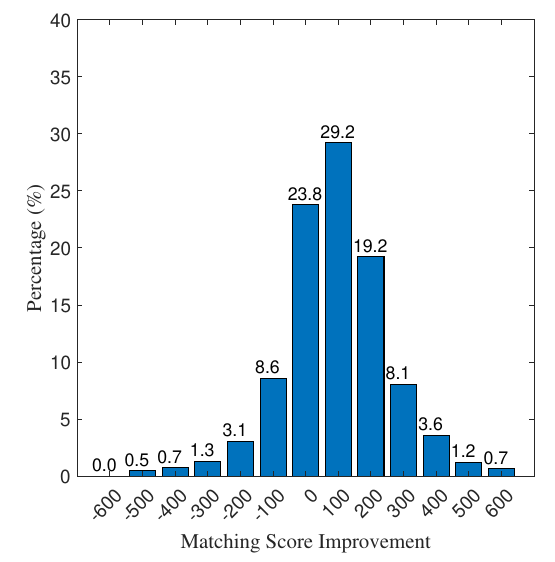}}
  \caption{The effect of different unfolding methods on genuine matching scores. (a) The VeriFinger matching score of unfolded fingerprints after general unfolding (X-axis) and specific unfolding (Y-axis). The red dotted line indicates that the scores obtained by the two methods are the same. (b)  Histogram of matching score improvement from general pose to specific pose.}
  \label{fig:regist}
\end{figure}

The purpose of 3D fingerprint unfolding is to improve the compatibility of 3D fingerprints and flat fingerprints, that is, to improve the matching score between them. We match the results of two unfolding algorithms with the same flat fingerprint, and the relative level of the matching score (VeriFinger \cite{VeriFinger} is used to calculate the matching score) can reflect the performance of unfolding algorithms.
Due to high fingerprint quality in the database, genuine match scores are much higher than impostor match scores and genuine matches and impostor matches are totally separable. Therefore, we use the improvement of genuine matching score as an indicator of unfolding quality.

Two unfolded fingerprints are cropped before computing VeriFinger matching scores according to the segmentation map of the real flat fingerprint to avoid the influence of redundant areas.
According to Fig. \ref{fig:regist}, we can see that most of the matching performance are improved after pose-specific unfolding and only a small part of the fingerprint matching scores  slightly drop. The curve shows that unfolding in a specific pose will have a better performance than unfolding directly. We analyzed the cases where pose-specific unfolding failed and found that in these cases, 3D pose estimation is inaccurate due to very small number of matching minutiae between 3D fingerprint and flat fingerprint.

\subsection{Deformation Field}

We estimate the deformation field between flat fingerprints and unfolded fingerprints with general pose and specific pose to directly measure the performance of the unfolding algorithm. We use the matching minutiae of the unfolded fingerprint and the flat fingerprint as control points and estimate the deformation field through thin plate spline interpolation. 

\begin{figure}[htb]
  \centering
    {\includegraphics[width=0.9\linewidth]{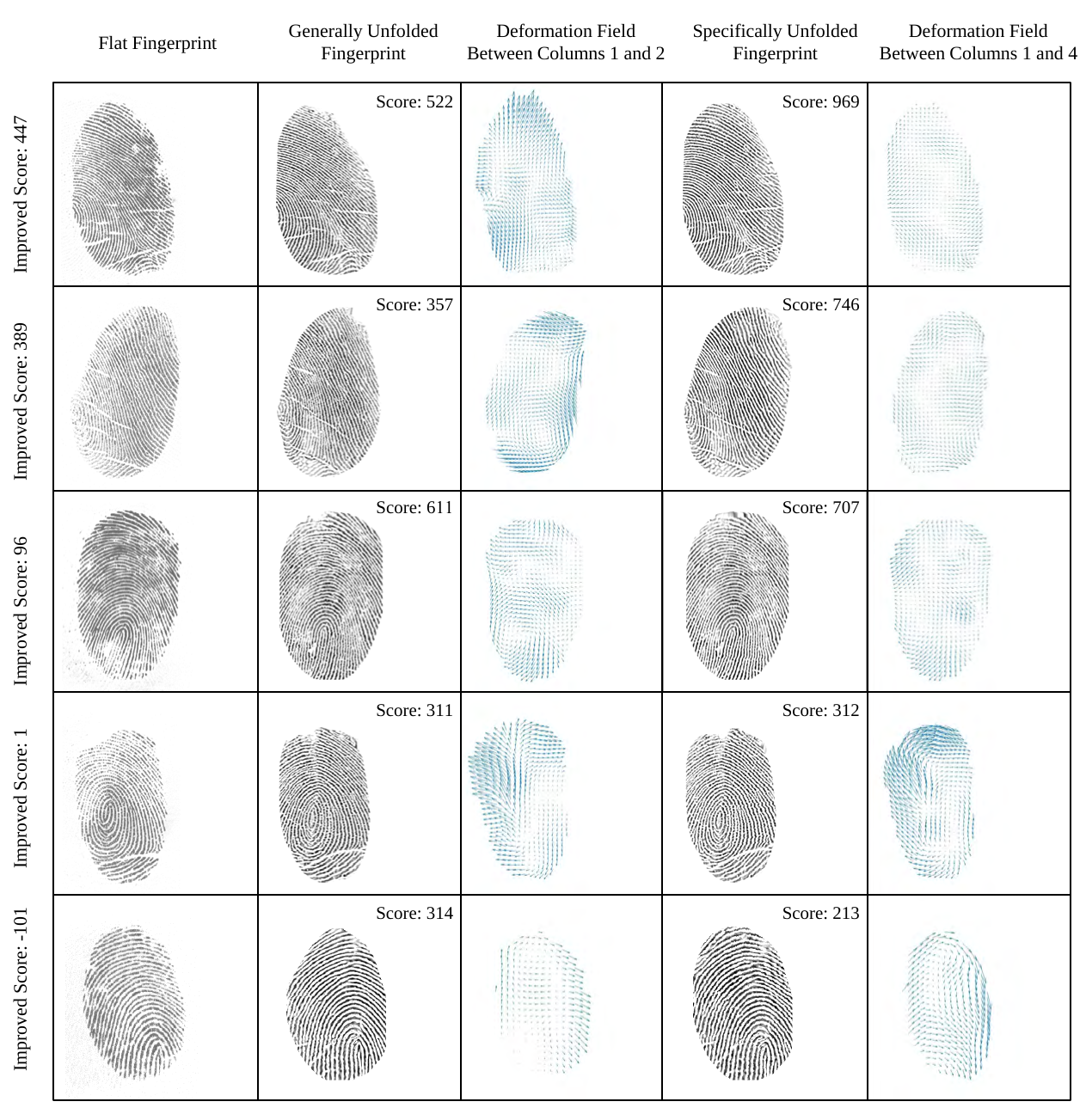}}
  \caption{The deformation fields and VeriFinger matching scores between flat fingerprints and unfolded fingerprints with general pose and specific pose. Longer arrows indicate larger deformation.}
  \label{fig:deformation}
\end{figure}

Five representative examples are given in Fig. \ref{fig:deformation}. Each row corresponds to a pair of 3D and 2D flat fingerprints.
These examples show that the improvement of the matching score is closely related to reduction of fingerprint distortion: when fingerprint distortion is greatly reduced, the matching score improves very much; when fingerprint distortion is almost unchanged, the matching score hardly improves. The comparison also proves that pose-specific unfolding does greatly reduce fingerprint deformation caused by different finger poses. It should be noted that the different finger poses are only part of the factors that cause deformation. Therefore, the deformation will only be relatively reduced but will not disappear completely by using the proposed unfolding algorithm.

\subsection{Efficiency}
The average times of major steps of the proposed unfolding algorithm are reported in Table \ref{tab:speed}. We ran the algorithm on a PC with 2.50 GHz CPU. All functions are implemented in MATLAB except for minutiae extraction which is performed using VeriFinger. 
The total time for the proposed unfolding algorithm is about 14 s. Since the fingerprint pairs have to be matched twice and the image size is large (900 pixel $\times$ 900 pixel), the feature extraction and matching part is time-consuming. Compared with general unfolding, our method requires additional steps for 3D pose estimation, 3D rigid transformation and specific unfolding, which take about 6.8 s.

\begin{table}
  \centering
  \caption{Speed of major steps of the proposed unfolding algorithm.}
  \begin{tabular}{ll}
    \toprule
      \qquad $\textbf{Step}$ \qquad \qquad & $\textbf{Time(s)}$ \qquad \\
    \midrule
       General unfolding \qquad \qquad &  3.08 \\
       Feature extraction \qquad \qquad &  2.17 \\
       3D pose estimation  \qquad \qquad &  2.23 \\
       3D rigid transformation \qquad \qquad &  1.44 \\
       Specific unfolding \qquad \qquad &  3.12 \\
       2D rigid transformation and cropping \qquad \qquad &  2.32 \\
    \bottomrule
  \end{tabular}
  \label{tab:speed}
\end{table}

\section{Conclusion} \label{sec:conclusion}
There are significant modal differences between 3D fingerprints and 2D fingerprints. To address this compatibility problem, existing algorithms focus on unfolding 3D fingerprints in the enrollment stage without considering the pose of the query fingerprint. 
In this paper, we propose a novel pose-specific 3D fingerprint unfolding algorithm. By unfolding the 3D fingerprint in a specific pose, the deformation is effectively reduced which improves the matching score with the real flat fingerprint. Meanwhile, our current method still has some shortcomings. The algorithm is slow and needs to be performed for each query fingerprint. Larger dataset should be collected to understand the performance on fingerprints of various poses and quality. In addition, the pressing force will also cause the deformation of the fingers, which are not considered. 
We plan to address the above limitations in the future.

\subsubsection*{Acknowledgments.} This work was supported in part by the National Natural Science Foundation of China under Grant 61976121.


\end{document}